%% file: paper_template.tex
\documentclass[conference]{IEEEtran}
\usepackage{newtxtext}
\usepackage{textcomp}
\usepackage{amsmath}
\usepackage{amsthm}
\theoremstyle{definition}
\newtheorem*{definition}{Definition}
\usepackage{booktabs}
\usepackage{graphicx}
\usepackage{tikz}
\usepackage{pgfplots}
\usepackage{pgfplotstable}
\pgfplotsset{compat=1.18}
\usepgfplotslibrary{groupplots,fillbetween}
\usepackage[most]{tcolorbox}
\definecolor{boxtitle}{RGB}{26,26,160}   % navy header
\definecolor{boxback}{RGB}{233,233,250}  % light lavender body

\newtcolorbox{claimbox}{
  colback=boxback,
  colframe=boxtitle,
  coltitle=white,
  fonttitle=\bfseries,
  title={Central Claim},
  boxrule=0.8pt,
  arc=2pt,
  left=5pt, right=5pt, top=4pt, bottom=4pt,
}

\newtcolorbox{implicationbox}{
  colback=boxback,
  colframe=boxtitle,
  coltitle=white,
  fonttitle=\bfseries,
  title={Implication for world models is structural},
  boxrule=0.8pt,
  arc=2pt,
  left=5pt, right=5pt, top=4pt, bottom=4pt,
}

\newtcolorbox{takeawaybox}[1][Takeaway]{
  colback=boxback,
  colframe=boxtitle,
  coltitle=white,
  fonttitle=\bfseries,
  title={#1},
  boxrule=0.8pt,
  arc=2pt,
  left=5pt, right=5pt, top=4pt, bottom=4pt,
}

\newcommand{\datadir}{.}
\input{tex/styles.tex}
\input{tex/headline_template.tex}

\usepackage[numbers]{natbib}
\usepackage{multicol}
\usepackage[bookmarks=true]{hyperref}

\begin{document}

\title{Imagined Rollouts Are Kinematic, Not Dynamic: \\A Diagnosis of Long-Horizon World-Model Failure}

% You will get a Paper-ID when submitting a pdf file to the conference system
%\author{Author Names Omitted for Anonymous Review. Robot World Model WS [Paper ID: 0005]}

\author{
\authorblockN{Finn Rasmus Sch\"afer\textsuperscript{1},
Korbinian Moller\textsuperscript{1},
Yuan Gao\textsuperscript{1},
Christian Oefinger\textsuperscript{1},\\
Sebastian Schmidt\textsuperscript{2}\,
and Johannes Betz\textsuperscript{1}}
\authorblockA{\textsuperscript{1}Autonomous Vehicle Systems Lab\\Technical University of Munich, Garching b. M\"unchen, Germany \\
Email: finn.schaefer@tum.de}
\authorblockA{\textsuperscript{2}Data Analytics and Machine Learning Group\\ Technical University of Munich, Garching b. M\"unchen, Germany}
}

% avoiding spaces at the end of the author lines is not a problem with
% conference papers because we don't use \thanks or \IEEEmembership

%Homer Simpson\authorrefmark{2},
%Montgomery Scott\authorrefmark{3} and
%Eldon Tyrell\authorrefmark{4}}
%Telephone: (800) 555--1212, Fax: (888) 555--1212}
%\authorblockA{\authorrefmark{4}Tyrell Inc., 123 Replicant Street, Los Angeles, California 90210--4321}}

\maketitle

\begin{abstract}
Long-horizon failure in world models is conventionally attributed to compounding error, a generic framing that does not distinguish what kind of error compounds. We propose a kinematic-vs-dynamic reframing: world models tend to imagine kinematically rather than dynamically. We operationalize this as the imagined Kinematic-Consistency Error, a per-step diagnostic that measures how far a rollout departs from a closed-form kinematic null, paired with a perturbation protocol that tests whether iKCE responds when physical conditions cross a regime boundary. We instantiate the diagnostic on a released DreamerV3 checkpoint trained on DMC walker-walk, where imagined iKCE runs roughly two orders of magnitude above that of matched real-physics rollouts. Across a friction sweep that crosses the gait-collapse boundary, the model's iKCE stays statistically flat even as the trained policy's reward collapses through the same range, providing the kinematic-not-dynamic signature. The diagnostic distinguishes kinematic from dynamic imagination at horizons longer than the embodiment's gait period.
\end{abstract}

\IEEEpeerreviewmaketitle

\section{Introduction}\label{sec:introduction}

World models have become a load-bearing component of recent embodied AI, serving as latent simulators for planning~\citep{ha2018worldmodels, hafner2019planet, hafner2023dreamerv3} and as generative environments for self-supervised learning~\citep{hu2022mile, hu2023gaia1}. A widely-noted failure mode is the deterioration of imagined rollouts over long horizons, conventionally attributed to \emph{compounding error}~\cite{janner2019mbpo}.
This framing is accurate but underspecified: it does not distinguish what kind of error compounds or which feature dimensions deteriorate. Across four recent
Observations in driving VLM/VLA and trajectory-prediction benchmarks indicate that the deterioration carries a specific structural signature that the compounding-error framing obscures.

We adopt the classical mechanics distinction between kinematic and dynamic motion. We define \emph{kinematic} as motion described purely through position, velocity, and acceleration time series, without invoking the forces or physical constraints that produced it. As \emph{dynamic}, we define motion that requires those constraints (e.g., mass, friction, contact) to be reproduced correctly.

\begin{claimbox}
\emph{Current world models imagine \textbf{kinematically} rather than \textbf{dynamically}:} they extrapolate position-velocity-acceleration trajectories that are internally consistent with linear kinematic update rules, but inconsistent with the physical constraints that produce real motion.
\end{claimbox}

Kinematic fallback is a third, structurally distinct account of long-horizon world-model failure, alongside the two positions that dominate the literature: \emph{predictable-representation engineering} (the Dreamer line~\citep{hafner2023dreamerv3}) and \emph{error-compounding bounds} (MBPO~\citep{janner2019mbpo}). Predictable-representation engineering attributes long-horizon reliability to stable, predictable latent representations and pursues it through normalization and balancing techniques; a single algorithm with fixed hyperparameters across 150+ tasks empirically validates the position, yet it is silent on \emph{what} those representations should contain. Error-compounding bounds derive an explicit quadratic-in-horizon bound on the gap between model-based and true returns under policy distribution shift, and respond pragmatically by limiting model trust to short branched rollouts from real states. A world model whose latent contains rich kinematic features but no dynamic features satisfies both: stable enough for Dreamer-line predictability, accurate enough inside the training regime for Janner-line short-rollout bounds, yet still biased toward kinematic continuation once conditioning pushes its rollouts across a physical-regime boundary. The three accounts are therefore not mutually exclusive but different layers of the same failure surface; because they predict distinct empirical signatures, the protocol of Section~\ref{sec:protocol} can target the kinematic-fallback layer specifically.

We make three contributions. We (i) \textbf{recast long-horizon world-model failure in kinematic-vs-dynamic terms}, distinguishing a structural-content layer from the variance-engineering and error-compounding layers studied in prior work; (ii) \textbf{introduce imagined kinematic-consistency error (iKCE) together with a conditioning-perturbation protocol} that operationalize this account as a falsifiable evaluation diagnostic; and (iii) \textbf{instantiate the diagnostic on an open-weight checkpoint} (DreamerV3 on DMC walker-walk), where it exhibits both predicted signatures of kinematic imagination, with controls ruling out the principal confounds. The two signatures are a kinematic-null residual $\sim$180$\times$ above matched physics at $T{=}16$, and statistical invariance of the imagined rollouts to a friction sweep that crosses the empirical gait-collapse boundary. The diagnostic signature is this regime-invariance, not the absolute iKCE magnitude: a trivially kinematic predictor would produce zero iKCE.

\section{Evidence}\label{sec:evidence}
Our diagnosis is motivated by four existing observations, each inconclusive alone and explained only in isolation by its original authors, but jointly forming a coherent structural signature.

\textbf{(i) Representational diagnostic on driving VLMs and VLAs.} \citet{schaefer2026egodynbench} present EgoDyn-Bench, a video-QA diagnostic that decouples physical reasoning from visual perception. (i), the weighted physics consistency rate (WPCR) saturates with a single static frame: rising from $\sim$20 with no visual input to $\sim$97 with one frame and remaining essentially flat as additional frames are added or temporally shuffled. (ii), reintroducing video to a text-only baseline recovers only $\sim$2.6pp on balanced accuracy under the best encoding, while text-only input already achieves 59.6\% BAcc. The authors characterize this as a ``functional decoupling between vision and language'': ego-motion understanding is derived almost exclusively from the language modality, with visual observations \textbf{Implication for world models is structural.} \textit{Contributing static context rather than temporal evidence. Imagined rollouts that depend on such encoders for motion-conditional features extrapolate from representations that under-encode the temporal dynamics they would need to imagine correctly.}

%The implication for world models is structural: imagined rollouts that depend on such encoders for motion-conditional features extrapolate from representations that under-encode the temporal dynamics they would need to imagine correctly.

\textbf{(ii) Sensor-degraded behavioral diagnostic.} \citet{priyadershi2026lostinfog} stress-test Alpamayo R1, a 10B-parameter driving VLA, across 1{,}996 scenarios under eight sensor perturbations. Under heavy Gaussian noise ($\sigma = 70$), the authors characterize the failure mode as one where the trajectory decoder ``fails via collapsing kinematic priors while the language branch continues producing coherent but safety-irrelevant explanations.'' 
\textbf{Independent observation.} \textit{This is an independent observation of the same kinematic-fallback failure mode, in a third-party VLA under naturalistic sensor degradation rather than controlled diagnostic stimuli.}
%This is an independent observation of the same kinematic-fallback failure mode, in a third-party VLA under naturalistic sensor degradation rather than controlled diagnostic stimuli.

\textbf{(iii) Open-loop trajectory-prediction baselines.} \citet{zhai2023admlp} train a 3-layer MLP that consumes only the ego vehicle's kinematic state and matches perception-based end-to-end planners on nuScenes open-loop L2 (0.29\,m vs.\ 0.37\,m for VAD-Base), with no camera, LiDAR, or HD-map input. The authors read this as a benchmark artifact, attributing it to the trajectory distribution of nuScenes and to a coarse collision-evaluation grid, and call for rethinking the open-loop evaluation scheme. We accept the empirical finding but read its significance differently.
\textbf{Third independent signature. }\textit{We read the same result as a third independent signature of kinematic imagination: when an ego-state-only predictor saturates the dominant open-loop metric, perception-based planners on this benchmark are not doing substantially more than kinematic extrapolation.}

%We read the same result as a third independent signature of kinematic imagination: when an ego-state-only predictor saturates the dominant open-loop metric, perception-based planners on this benchmark are not doing substantially more than kinematic extrapolation.

%\textbf{(iv) Physics-consistency scoring on fine-tuned VLAs.} \citet{gao2026stylevla} introduce a Kinematic Consistency Error (KCE) that quantifies physical inconsistency in a generated trajectory by checking each predicted next position against a closed-form kinematic extrapolation of the current state, and use the same expression as a training-time loss to supervise a fine-tuned 4B-parameter VLA. On their style-conditioned driving benchmark, KCE varies substantially across the model lineup (Gemini-3-Pro at 0.06-0.11\,m, fine-tuned 4B/7B models at 0.08--0.12\,m), with no monotonic relationship to model scale or modality. Strong, generalist models can be kinematically inconsistent, and this inconsistency can be detected with a lightweight per-step metric.
\textbf{(iv) Physics-consistency scoring on fine-tuned VLAs.} \citet{gao2026stylevla} introduce a Kinematic Consistency Error (KCE) that scores a trajectory by checking each predicted next position against a closed-form kinematic extrapolation of the current state, and reuse it as a training loss for a fine-tuned 4B VLA. On their style-conditioned benchmark, KCE shows no monotonic relationship to model scale or modality: the strongest generalist (Gemini-3-Pro, 0.06--0.11\,m) and the smaller fine-tuned models (0.08--0.12\,m) overlap, with no ordering by parameter count or sensor richness. \textbf{A data/training deficit, not a capacity one. }
\textit{A deficit that is invariant to model scale and modality is unlikely to reflect a capacity limitation. It points instead to the training signal and data distribution rather than to model size.}

\textbf{Conclusion.} The four observations are concentrated in the driving and VLA setting but employ heterogeneous methodological registers, from representational probing to behavioral perturbation to physics-consistency scoring, and converge on a single structural deficit: the learned representation is dominated by kinematic features, and the dynamic features required for physical-regime-conditional behavior are systematically under-represented.

\section{Diagnostic Protocol}\label{sec:protocol}
\subsection{iKCE: Imagined Kinematic Consistency Error}
  \label{sec:ikce}
% INTENT: define iKCE; flag it's a repurposed training loss
% (StyleVLA KCE / PIKC); critical-interpretation note that LOW
% iKCE is BAD (signals kinematic imagination); one display equation.
% CITATIONS: \cite{stylevla2026, egodyn2026, mbpo2019}
\begin{definition}[Imagined Kinematic-Consistency Error]
For an imagined rollout $\{\hat{x}^{\text{WM}}_t\}_{t=0}^{T}$ produced by a world model, with $\hat{x}_t$ a chosen kinematic state vector (e.g.\ $[x, y, v, a, \theta]^\top$), the \emph{imagined kinematic-consistency error} is
\begin{equation}
\label{eq:ikce}
  \mathrm{iKCE} \;\doteq\; \frac{1}{T}\sum_{t=0}^{T-1}
    \left\|\,\hat{x}^{\text{WM}}_{t+1} - \mathrm{kin}\!\left(\hat{x}^{\text{WM}}_t\right)\,\right\|^{2},
\end{equation}
where $\mathrm{kin}(\cdot)$ is any closed-form kinematic predictor (e.g., constant-velocity or constant-acceleration) chosen to match the WM's underlying embodiment and output space.
\end{definition}

Equation~\ref{eq:ikce} follows the mathematical form of the kinematic-consistency loss in \citet{gao2026stylevla}, which supervises a fine-tuned VLA at training time. We repurpose it as a test-time diagnostic on imagined rollouts (the prefixed ``i'' denotes \emph{imagined}), measuring kinematic inconsistency rather than reducing it. We repurpose it as an \emph{evaluation diagnostic}: applied at test time to imagined rollouts, iKCE measures how far each predicted next state \emph{departs} from the kinematic extrapolation of its predecessor.

Counter-intuitively, \textbf{low iKCE does not indicate dynamic imagination}: a world model with near-zero iKCE predicts, by construction, next states that coincide with the kinematic continuation of their predecessors: it imagines kinematically. The signature of dynamic imagination is the opposite: iKCE positive, growing with horizon, and responsive to physical-regime conditioning (friction transients, contact events, regime-boundary crossings). iKCE is therefore necessary but not sufficient to certify dynamic imagination.

\subsection{Conditioning perturbations}
\label{sec:perturbation}
% INTENT: turn static iKCE into dose-response; two diagnostic curves
% (regime-boundary spike, monotone-consistency); position as the
% closest defensible analog to EgoDyn-Bench WPCR.
Static iKCE measures internal kinematic self-consistency along a single rollout. To turn this into a diagnostic that \emph{separates} kinematic from dynamic imagination, we drive iKCE through a dose-response curve over the conditioning state. For each base rollout, the world model generates $K$ imagined rollouts under controlled perturbations of physically meaningful conditioning parameters, initial velocity $v_0$, friction coefficient $\mu$, or lateral-acceleration limit $a_{\text{lat,max}}$ for driving, terrain compliance, or payload for legged locomotion, analogous to physically-grounded parameters for other embodiments. The perturbation set is embodiment-specific. The protocol is not.

Two diagnostic signatures emerge from the resulting $\{\text{iKCE}_k\}_{k=1}^{K}$ ensemble. First, the \emph{shape} of $\text{iKCE}(\|\Delta\|)$ as a function of perturbation magnitude: a kinematic imaginer produces a curve that scales smoothly with $\|\Delta\|$ regardless of physical regime, because it is extrapolating the same linear update structure in every case. iKCE measures per-step kinematic-null deviation. Physical-regime perturbations whose effects are slow relative to the per-step timescale (e.g., friction-driven slipping in legged locomotion, which accumulates over multiple footfalls) are not visible at horizons shorter than their characteristic accumulation time. The diagnostic protocol should be applied at horizons longer than the embodiment's gait period: $\sim$25 ms $\times$ 64 steps $\approx$ 1.6 s is sufficient for walker-class locomotion. Driving trajectories are usually planned over longer horizons. Second, \emph{rollout-pair monotone consistency}: physics predicts monotone responses to specific perturbation pairs (higher initial velocity implies longer stopping distance under braking; heavier payload implies slower acceleration), and for each such pair $(\Delta_a, \Delta_b)$ we check whether the imagined rollouts respect the monotonicity. A kinematic world model trivially respects \emph{linear} monotonicities but fails \emph{physical-regime-conditional} ones, where the monotonicity only holds above or below a physical threshold.

This is the closest defensible analog to EgoDyn-Bench's weighted pairwise consistency rate~\citep{schaefer2026egodynbench}: rather than checking answer consistency across tagged question pairs on a single observation, we check rollout consistency across controlled conditioning perturbations on a single base scenario. The diagnostic is reframed rather than contrived, and produces falsifiable curves on any embodiment whose state admits a kinematic predictor.

\section{Experiments and Results}\label{sec:experiments}
The experimental setup is documented in Appendix~\ref{app:setup}.
\textbf{Hypothesis 1 (H1): iKCE is non-degenerate on a published checkpoint.}
At both measurement horizons, the trained DreamerV3 walker-walk world model produces an imagined iKCE at least an order of magnitude above matched policy-driven real-physics rollouts on the same conditioning (see Table~\ref{tab:h1}: $\sim$180$\times$ at $T{=}16$, $\sim$30$\times$ at $T{=}64$). The narrowing at the longer horizon is consistent with the per-step dilution noted in §V, limitation~(iii): the WM's smooth post-transient tail averages down the integrated metric. iKCE is therefore non-degenerate at both horizons (see the trivial-WM scale anchor
in Appendix~\ref{app:trivial-wm} for the analytic lower bound). The WM's imagination carries a substantial residual against any constant-velocity null, with the H1 being a lower bound on the magnitude separation.

\begin{table}[t]
\centering\small
\caption{iKCE on the $(z, v_z)$ view at $\mu{=}1.0$ baseline, $K{=}20$ rollouts, 95\% bootstrap CIs. WM iKCE exceeds physics by at least an order of magnitude at both measurement horizons. The ratio narrows at $T{=}64$ due to per-step dilution from the WM's smooth long-horizon tail (see §V, limitation~(iii)).}
\label{tab:h1}
\begin{tabular}{lcc}
\toprule
Source & Mean iKCE & 95\% CI \\
\midrule
\multicolumn{3}{l}{\emph{Horizon $T{=}16$}} \\
Real physics (matched policy) & $4.2{\times}10^{-5}$ & $[3.2,\, 5.3]{\times}10^{-5}$ \\
DreamerV3 WM (imagined)       & $7.7{\times}10^{-3}$ & $[5.2,\, 10.3]{\times}10^{-3}$ \\
Ratio (WM / physics)          & $\sim 180\times$ & --- \\
\midrule
\multicolumn{3}{l}{\emph{Horizon $T{=}64$}} \\
Real physics (matched policy) & $8.6{\times}10^{-5}$ & $[6.4,\, 11.0]{\times}10^{-5}$ \\
DreamerV3 WM (imagined)       & $2.6{\times}10^{-3}$ & $[2.0,\, 3.2]{\times}10^{-3}$ \\
Ratio (WM / physics)          & $\sim 30\times$ & --- \\
\bottomrule
\end{tabular}
\end{table}

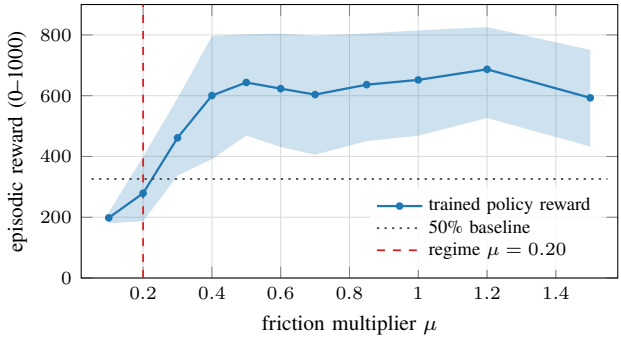
\begin{figure}[t]
  \centering
  \input{tex/regime_curve.tex}
  \caption{\textbf{Empirical regime boundary.} Mean episodic reward of the trained DreamerV3 walker policy across friction ($K{=}10$, 95\% bootstrap CI). The regime boundary at $\mu{=}0.20$ (red dashed) is the friction at which mean reward first drops below 50\% of the $\mu{=}1.0$ baseline (dotted). This boundary anchors the dashed reference line in Fig.~\ref{fig:headline}.}
  \label{fig:regime}
\end{figure}

\textbf{Hypothesis 2 (H2): WM imagination is friction-invariant across a regime boundary.}
We sweep surface friction across 13 magnitudes in $[0.1, 1.7]$, spanning the regime boundary at $\mu{=}0.20$ where the trained gait first drops below 50\% of baseline episodic reward (empirically determined, see Fig.~\ref{fig:regime}). For each $\mu$, we compute iKCE on (a) real-physics rollouts under the trained actor and (b) WM-imagined rollouts conditioned on the first 5 perturbed observations.

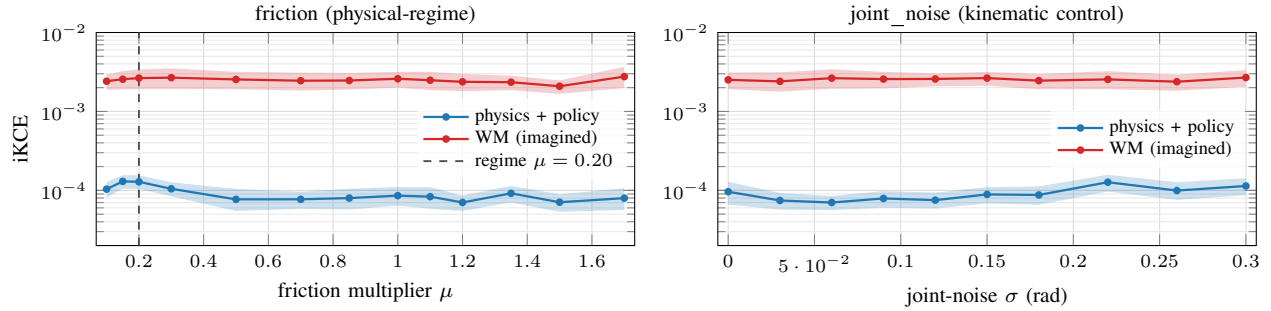
\begin{figure*}[t]
  \centering
  \input{tex/headline_identity_h64.tex}
  \caption{\textbf{iKCE diverges in physics, stays flat in imagination.}
           Identity kinematic view $(z, v_z)$ at horizon $T = 64$. \emph{Left:} friction sweep $\mu \in [0.1, 1.7]$ (physical regime axis). Real-physics rollouts under the trained policy (blue) show modestly elevated iKCE in the low-$\mu$ region near the empirical regime boundary ($\mu{=}0.20$, dashed line (see Fig.~\ref{fig:regime})). WM-imagined rollouts (red) are statistically flat across the entire $17\times$ sweep. \emph{Right:} joint-noise sweep $\sigma \in [0, 0.3]$\,rad (kinematic control axis). Both channels respond similarly, confirming that the WM is not insensitive to all perturbations, only to dynamic ones. Shaded bands: 95\% bootstrap CI from $K=20$ rollouts per cell.}
  \label{fig:headline}
\end{figure*}

At $T{=}64$, WM iKCE across the sweep is statistically flat (max/min spread $1.32\times$, 95\% CIs overlap at every $\mu$, see Fig.~\ref{fig:headline}). A log-log regression on the same sweep makes this falsifiable (Appendix~\ref{app:setup}): the WM slope's 95\% bootstrap CI ($\beta_{\mathrm{WM}}=-0.009$, $[-0.096, +0.082]$) contains zero, while the physics slope's $\left(\beta_{\mathrm{phys}}=-0.220 \text{, } [-0.301, -0.142]\right)$ does not. The trained policy's reward, by contrast, collapses through the same range (from $\sim$650 at $\mu{=}0.5$ to $\sim$200 at $\mu{=}0.10$), a real behavioral regime change to which the WM's imagined rollouts are blind. Real-physics iKCE under the trained actor shows more cell-to-cell variability than the WM, with elevated values in the low-$\mu$ region (means $1.04$-$1.30 \times 10^{-4}$ across $\mu \in [0.1, 0.3]$ vs.\ $0.70$--$0.92 \times 10^{-4}$ across $\mu \in [0.5, 1.7]$). This confirms that the iKCE metric is not degenerate, while the WM's invariance to the same perturbation provides the kinematic, not dynamic, signature. The structural difference is reinforced by appendix controls: a per-step decomposition (Fig.~\ref{fig:perstep}) shows that the physics and WM residuals differ qualitatively in temporal structure, not just in magnitude, and the same H2 signature reappears under a richer kinematic slice (Fig.~\ref{fig:gait}). At the shorter $T{=}16$ horizon, neither side shows friction sensitivity. The dynamic signature emerges only as slip accumulates into per-step deviation at the longer horizon (quantified in Appendix Fig.~\ref{fig:horizon-sweep}). Controls supporting H2 are reported in detail in the appendix: per-step structure under the actor ablation (Fig.~\ref{fig:perstep_a64}), robustness to the kinematic-state choice (Fig.~\ref{fig:gait}), and a joint-noise positive control (Fig.~\ref{fig:headline}, right panel) confirming the WM responds to kinematic-axis perturbations.

\textbf{Actor training horizon ablation.} An actor-training-horizon ablation (see Appendix Fig.~\ref{fig:headline_a64}) rules out the most concerning confound: retraining at imag horizon\,=\,64 produces identical WM iKCE friction spread (1.32$\times$), confirming H2 is not an artifact of the default actor's 15-step training horizon. 

\textbf{Domain-randomization control.} A domain-randomization control (Appendix~\ref{app:dr-control}, Fig.~\ref{fig:headline_dr}) bounds the policy-OOD confound: under a policy trained with $\mu \sim \mathcal{U}(0.1, 1.7)$, the WM slope still contains zero ($\beta^{\mathrm{DR}}_{\mathrm{WM}} = -0.026$) and the physics slope still excludes it ($\beta^{\mathrm{DR}}_{\mathrm{phys}} = -0.114$), so the kinematic-not-dynamic contrast survives with the policy in-distribution at every swept friction.

\section{Discussion \& open directions}\label{sec:discussion}
iKCE is a per-step kinematic null fit integrated over a horizon. It diagnoses kinematic imagination by the absence of regime sensitivity, not by the presence of dynamic prediction quality. A WM that achieves low iKCE everywhere has been correctly identified as kinematic by our protocol, but has not been certified as a useful predictor. A high-iKCE WM with friction sensitivity has been certified as dynamic but not as accurate. The diagnostic is structural, not predictive, by design.

\textbf{A downstream behavioral prediction.} If imagined rollouts are kinematically structured but not dynamically faithful, then policy gradients propagated through long imagined rollouts optimize the actor against a trajectory distribution that diverges from real dynamics in directions iKCE itself does not capture (rotational drift, contact-event timing, accumulated absolute-state error). Long-horizon actor training should therefore be unstable and yield a weaker deployed policy. The $h{=}64$ ablation is consistent with this prediction: under matched hyperparameters, the long-horizon actor converged to $\sim$400 episodic reward versus $\sim$955 for the default $h{=}15$ checkpoint, and exhibited training instability throughout. We do not claim a causal link. Long-horizon Dreamer training is known to be sensitive to multiple factors, but the observation is what one would predict from the kinematic-not-dynamic hypothesis, and pre-empts the natural counterfactual that scaling the actor's imagination horizon would have closed the gap.

\textbf{Limitations of the present measurement.}
Several limitations bound the result. (i) The empirical result rests on a single embodiment (DMC walker-walk, a 2D 9-DOF system) restricted to the $(z, v_z)$ sub-slice, and on a single open-weight WM family. Extending the flatness test for H2 to quadruped, humanoid, and driving embodiments, as well as to other WM families such as GAIA-1~\cite{hu2023gaia1} or R2Dreamer~\cite{morihira2026r2dreamerredundancyreducedworldmodels}, would further broaden the evidence base. (ii)~The policy-OOD confound of the physics-side signature is bounded, not eliminated, by the domain-randomization control (Appendix~\ref{app:dr-control}): the H2 contrast survives under a policy in-distribution at every swept $\mu$, while the partition of the original low-$\mu$ elevation holds at the point-estimate level only. Two residuals remain: the DR policy adapts its gait to friction in closed loop, so the physics response is not policy-free; and imagined rollouts condition on only five observations (under one gait period), so the WM-side flatness is informative only up to the regime evidence the prefix can carry. (iii)~Per-step displacement decays over the rollout horizon, so the WM's low long-horizon iKCE reflects in part reduced motion magnitude rather than purely cleaner kinematic imagination.

\textbf{Open directions.} Embodiment extension to quadruped-walk would test the diagnostic on richer contact dynamics than the planar walker provides. Fixed open-loop action sequences, applied identically across physics and WM rollouts, would remove the closed-loop policy adaptation left in place by the domain-randomization control. An explicit-conditioning experiment, in which friction or contact indicators are appended to the WM's observation, together with a conditioning-prefix-length sweep (5--64 observed steps), would distinguish representational absence from architectural insensitivity -- and from regime evidence the prefix simply cannot carry. A driving-WM cross-anchor on a model exposing an ego-pose head (Vista~\cite{gao2024vistageneralizabledrivingworld}, DriveDreamer~\cite{wang2023drivedreamerrealworlddrivenworldmodels}) would connect the measurement to the autonomous-driving evidence of \S\ref{sec:evidence}.

%\clearpage

% TODO: Camera Ready Version
%\section*{Acknowledgments}

%% Use plainnat to work nicely with natbib. 

\bibliographystyle{plainnat}
\bibliography{references}

\clearpage

\appendix
\section{Experimental Setup}\label{app:setup}

This appendix documents (i) the experimental configuration (Table~\ref{tab:experimental-setup-test}), (ii) the methodological
details behind the headline numbers (\S\ref{app:regime-bound}-\ref{app:trivial-wm}: regime-boundary determination, the flatness and horizon-emergence tests, and the trivial-WM scale anchor), (iii) the controls supporting H2 (\S\ref{app:actor-ablation}-\ref{app:joint-noise}: actor-training-horizon ablation, domain-randomization control, per-step structure decomposition, per-step structure under the actor ablation, robustness to the kinematic-state choice, and the joint-noise positive control), and (iv) implementation specifics needed to reproduce the measurement (\S\ref{app:code}-\ref{app:impl}). Code and data will be released upon acceptance.

\begin{table}[htbp]
\centering\small
\caption{Experimental setup. Horizons $h{=}16$ and $h{=}64$ refer
to the iKCE rollout length, evaluated on the same trained policy.}
\label{tab:experimental-setup-test}
\begin{tabular}{@{}lp{0.62\columnwidth}@{}}
\toprule
Field & Value \\
\midrule
World model        & DreamerV3 (NM512 PyTorch port, commit \texttt{6ef8646}) \\
Task               & DMC walker-walk, \texttt{dmc\_proprio} config \\
Training           & 1M env steps, seed 0, RTX 5090 \\
Final reward       & $955 \pm 30$ (mean over last 100k steps) \\
Evaluation policy  & Trained actor (same checkpoint on both physics and WM sides) \\
Backend            & \texttt{dm\_control} 1.0.20, \texttt{mujoco} 3.1.6 \\
$K$                & 20 rollouts per perturbation cell \\
Kinematic spec     & $(z, v_z)$ root-vertical-motion (1D) \\
Extrapolation      & constant velocity \\
\bottomrule
\end{tabular}
\end{table}

\subsection{Methodological Details}
\subsubsection{Regime-boundary determination.}\label{app:regime-bound}
The boundary $\mu{=}0.20$ referenced in Fig.~\ref{fig:headline} (dashed line) and Fig.~\ref{fig:regime} is not chosen a priori. It is determined empirically from the policy's reward collapse. We roll out the trained actor for $K{=}10$ episodes at each of 12 friction multipliers $\mu \in [0.1, 1.5]$, compute the mean episodic reward and a $95\%$ bootstrap CI per cell, and define the regime boundary as the \emph{largest} $\mu$ at which mean reward has dropped below $50\%$ of its $\mu{=}1.0$ baseline. The $\mu{=}1.0$ mean over this sweep is $\sim$650 ($K{=}10$ episodes), giving a threshold of $\sim$325. On our checkpoint, the boundary lies at $\mu{=}0.20$. (The $955 \pm 30$ final reward in Table~\ref{tab:experimental-setup-test} averages over the last 100k training steps and is not directly comparable.)

\subsubsection{Flatness test for H2.}\label{app:flatness}
We make the ``statistically flat'' claim falsifiable by regressing $\log(\text{iKCE})$ on $\log(\mu)$ across the $T{=}64$ friction sweep, with each of the $K{=}20$ rollouts at each of the 13 $\mu$ values contributing one observation ($n{=}260$ per seed). To rule out a seed-dependent artifact, we repeat the regression for three independently-trained DreamerV3 walker-walk checkpoints (seeds $\{0, 1, 2\}$, matched hyperparameters and step budget; see Appendix~\ref{app:setup}). A 95\% percentile bootstrap (1000 resamples over rollouts) on each slope gives: $\beta_{\mathrm{WM}}^{(0)} = -0.009$, CI $[-0.096, +0.082]$; $\beta_{\mathrm{WM}}^{(1)} = +0.031$, CI $[-0.072, +0.129]$; $\beta_{\mathrm{WM}}^{(2)} = +0.038$, CI $[-0.039, +0.123]$, all three WM slope CIs contain zero, so H2's flatness claim survives a falsifiable statistical test across seeds, not just on the original training run. The same procedure on the physics-side sweep (seed 0 only, as physics is not a learned model) gives $\beta_{\mathrm{phys}} = -0.220$, CI $[-0.301, -0.142]$, comfortably excluding zero. The low-$\mu$ elevation reported in \S\ref{sec:experiments} is statistically real. The quantitative form of the kinematic-not-dynamic signature is therefore: across three seeds, $|\beta_{\mathrm{WM}}|$ is bounded above by $\sim$0.13 (one decade of $\mu$ changes WM iKCE by at most $\sim$13\%), while $|\beta_{\mathrm{phys}}| \approx 0.22$ ($\sim$25\% per decade), with non-overlapping confidence intervals.

\subsubsection{Horizon-emergence test.}\label{app:horizon-emergence}
\begin{figure}[!ht]
  \centering
  \input{tex/horizon_sweep.tex}
  \caption{\textbf{Horizon-emergence test.} Slope $\beta = \partial \log(\text{iKCE}) / \partial \log(\mu)$ of the friction sweep at four measurement horizons $T \in \{8, 16, 32, 64\}$, computed by re-integrating the saved $T{=}64$ per-step iKCE traces (no new rollouts). Physics slope (blue) grows in magnitude with $T$ and crosses out of the CI-contains-zero region by $T{=}64$. WM slope (red) is statistically indistinguishable from zero at every horizon. Dashed line marks the H2 flatness target ($\beta = 0$). Shaded bands: 95\% percentile bootstrap CI over 1000 resamples at the rollout level.}
  \label{fig:horizon-sweep}
\end{figure}
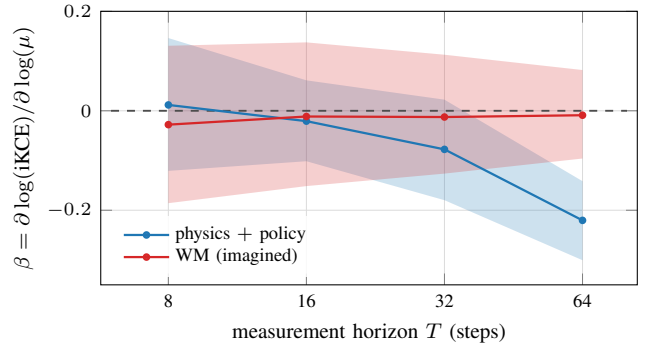

Section~\ref{sec:perturbation} argues that the diagnostic should be applied at horizons longer than the embodiment's gait period. We sharpen this claim quantitatively by repeating the flatness regression of the preceding paragraph at four sub-horizons $T \in \{8, 16, 32, 64\}$, re-integrating each rollout's saved per-step iKCE trace from the existing $T{=}64$ sweep (no new rollouts, see Fig.~\ref{fig:horizon-sweep}). The dynamic signature in physics emerges with horizon: the slope $\beta_{\mathrm{phys}}$ grows in magnitude from $+0.012$ ($\mathrm{CI}~[-0.121, +0.146]$) at $T{=}8$ to $-0.221$ ($\mathrm{CI}~[-0.301, -0.142]$) at $T{=}64$, crossing out of the CI-contains-zero region between $T{=}32$ and $T{=}64$~-- consistent with friction effects accumulating over multiple footfalls before becoming detectable in the per-step kinematic-null residual. The WM-side slope $\beta_{\mathrm{WM}}$ is statistically indistinguishable from zero at every horizon tested ($-0.028$, $-0.012$, $-0.013$, $-0.009$ at $T{=}8, 16, 32, 64$, with CIs of width $\le 0.32$ all straddling zero). The contrast is the H2-emergence claim stated quantitatively: the dynamic signature emerges with horizon in physics but not in the WM. Note that long-horizon iKCE in both channels is in part dilated by reduced per-step motion magnitude (\S\ref{sec:discussion}, limitation~(iii)). The present result is robust to that effect because the WM-physics contrast \emph{widens} with horizon rather than shrinks, which is the opposite of what a horizon-degenerate metric would produce.

\subsubsection{Trivial-WM scale anchor.}\label{app:trivial-wm} 
The iKCE scale has an analytic lower bound that anchors the magnitudes in Table~\ref{tab:h1}. A trivial ``WM'' that imagines by applying the kinematic predictor to its own current state, $\hat{x}^{\text{WM}}_{t+1} = \mathrm{kin}(\hat{x}^{\text{WM}}_t)$, produces $\text{iKCE} = 0$ by construction: each predicted next state is identically the kinematic continuation of its predecessor, so every residual in Eq.~\ref{eq:ikce} is zero. The measured ordering on walker-walk is therefore
\[
  \underbrace{0}_{\text{trivial kinematic}}
  \;<\;
  \underbrace{4.2 \times 10^{-5}}_{\text{matched real physics}}
  \;\ll\;
  \underbrace{7.7 \times 10^{-3}}_{\text{DreamerV3 WM}}
  \quad (T{=}16).
\]
The WM lies further from the trivial-kinematic baseline than real physics does, ruling out a naive reading in which ``imagining kinematically'' would imply small absolute iKCE. The diagnostic signature, per Section~\ref{sec:protocol}, is friction-invariance under perturbation, not low absolute magnitude. The order holds for $T{=}64$ as well.

\subsection{Controls}
\subsubsection{Actor-training-horizon control.}\label{app:actor-ablation} 
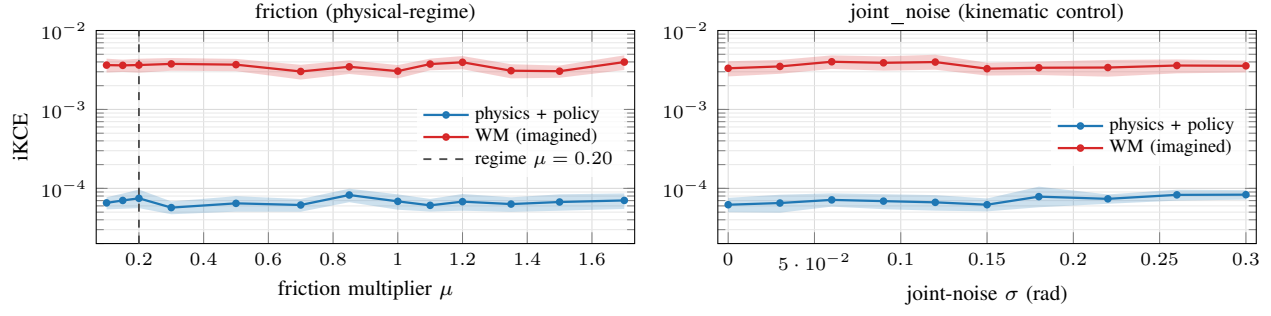
\begin{figure*}[!ht]
  \centering
  \input{tex/headline_identity_a64.tex}
  \caption{\textbf{Actor-training-horizon ablation.} Identity view $(z, v_z)$ at $T{=}64$, with the actor retrained at
           \texttt{imag horizon}~$=64$ (matching the measurement horizon). WM iKCE friction spread is identical to the
           default-actor headline ($1.32\times$ in both cases), confirming the friction-insensitivity of imagined rollouts
           is not an artifact of the default actor's $h{=}15$ training horizon. Shaded bands: 95\% bootstrap CI from $K=20$
           rollouts per cell.}
  \label{fig:headline_a64}
\end{figure*}

A natural concern is that the WM's friction-invariance at $T{=}64$ reflects the default actor operating out-of-distribution from its training horizon (\texttt{imag horizon}~$=15$) rather than a structural property of WM imagination. To rule this out, we retrain an identical checkpoint at \texttt{imag horizon}~$=64$, matching the measurement horizon, with the same seed, hyperparameters, and total training budget. Fig.~\ref{fig:headline_a64} shows the result: WM iKCE friction spread under the $h{=}64$-trained actor is identical to the default-actor headline ($1.32\times$ in both cases, CIs overlapping at every $\mu$), confirming that friction-invariance is not an artifact of actor training horizon.

\subsubsection{Domain-randomization control.}\label{app:dr-control}
Limitation~(ii) in \S\ref{sec:discussion} names the most consequential confound of the physics-side H2 signature: the evaluation policy acted only at $\mu{=}1.0$ during training, so the elevated low-$\mu$ physics iKCE may reflect an in-distribution policy slipping under out-of-distribution friction rather than a genuine friction response of the contact dynamics. To quantify this confound, we train a fourth, otherwise identical DreamerV3 checkpoint with per-episode domain randomization of friction, $\mu \sim \mathcal{U}(0.1, 1.7)$ drawn at every episode reset (matched hyperparameters and step budget; evaluation reward $930 \pm 36$ across the full sweep range). Both the DR policy and the DR world model are therefore in-distribution at every friction value of the H2 sweep. The DR policy exhibits no reward collapse anywhere in the tested range, so the regime boundary of Fig.~\ref{fig:regime} is a property of the fixed-$\mu$ policy and does not transfer to this control.
 
\begin{figure*}[!ht]
  \centering
  \input{tex/headline_identity_dr.tex}
  \caption{\textbf{Domain-randomization control.} Identity view $(z, v_z)$ at $T{=}64$ with friction domain-randomized at training time ($\mu \sim \mathcal{U}(0.1, 1.7)$ per episode). Left: friction sweep $\mu \in [0.1, 1.7]$ (physical-regime axis). WM-imagined rollouts remain statistically flat despite the world model having been trained on the full friction range; physics-side iKCE under the DR policy retains a strictly negative slope, with the point estimate at roughly half the default-policy magnitude. Right: joint-noise sweep $\sigma \in [0, 0.3]$\,rad (kinematic control axis). The DR-trained WM responds to kinematic-axis perturbations, replicating the positive control of Fig.~\ref{fig:headline} under the DR checkpoint. Shaded bands: 95\% bootstrap CI from $K{=}20$ rollouts per cell.}
  \label{fig:headline_dr}
\end{figure*}
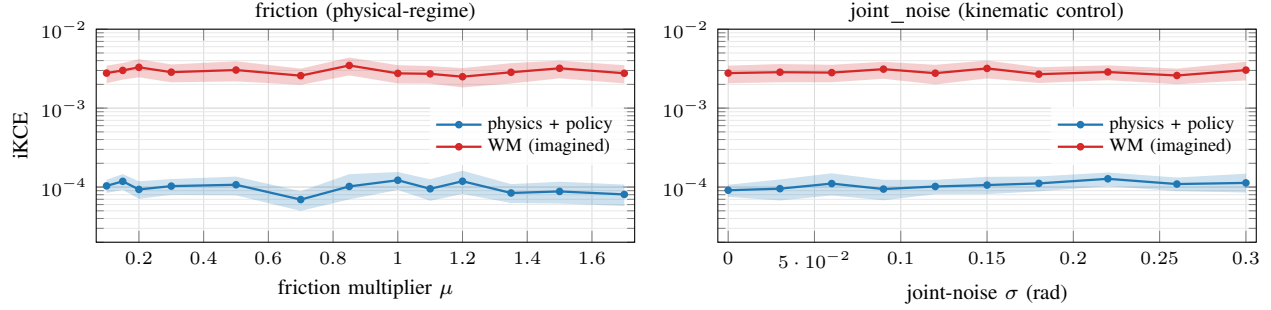

Table~\ref{tab:dr-flatness} reports the flatness regression of Appendix~\ref{app:flatness} under the DR checkpoint, alongside the fixed-$\mu$ results; Fig.~\ref{fig:headline_dr} shows the underlying sweeps. \emph{WM side:} the DR-trained world model is statistically flat ($\beta^{\mathrm{DR}}_{\mathrm{WM}} = -0.026$, CI $[-0.123, +0.076]$), indistinguishable from the three fixed-$\mu$ seeds. This closes a data-coverage loophole in the seed-level flatness test: a world model trained only at $\mu{=}1.0$ has never observed friction variation and cannot have learned friction-conditional latent dynamics, so its flatness is partially guaranteed by construction. The DR world model was trained on transitions spanning the full sweep range, could in principle infer the friction regime from its conditioning prefix and imagine regime-conditional rollouts -- and it remains friction-invariant (Fig.~\ref{fig:headline_dr}, left). The flatness signature is therefore not an artifact of the training-time friction distribution.
 
\emph{Physics side:} under the DR policy, the physics slope remains strictly negative ($\beta^{\mathrm{DR}}_{\mathrm{phys}} = -0.114$, CI $[-0.201, -0.024]$), at roughly half the default-policy magnitude ($-0.220$). Read at the point-estimate level, this partitions the original low-$\mu$ elevation into comparable parts: about half attributable to an out-of-distribution policy slipping, about half a genuine friction response of the contact dynamics that persists when the policy is in-distribution everywhere. We state this partition at the point-estimate level only: a percentile bootstrap on the difference of the two physics slopes does not resolve it at $K{=}20$ ($\beta^{\mathrm{default}}_{\mathrm{phys}} -\beta^{\mathrm{DR}}_{\mathrm{phys}} = -0.107$, CI $[-0.221, +0.004]$).

Under matched DR conditions on both sides, the H2 contrast retains its falsifiable form: the physics slope's CI excludes zero while the WM slope's contains it. The slope difference itself is directionally consistent but not resolved at this sample size ($\beta^{\mathrm{DR}}_{\mathrm{phys}} - \beta^{\mathrm{DR}}_{\mathrm{WM}} = -0.087$, CI $[-0.211, +0.049]$).
 
This control addresses the training-distribution side of limitation~(ii). Two residual caveats remain, both deferred to the open directions: the DR policy still adapts its gait to friction in closed loop, so the fully policy-free variant (fixed open-loop action sequences applied identically to both channels) remains the cleaner disambiguation; and the imagined rollouts condition on only five observed steps ($125$\,ms, under one gait period), which bounds how much regime evidence even a dynamically capable imaginer could extract from the prefix.
 
\begin{table}[htbp]
\centering\small
\caption{Domain-randomization control: flatness regression $\beta = \partial \log(\text{iKCE}) / \partial \log(\mu)$ at $T{=}64$, 95\% percentile bootstrap CIs (1000 resamples at the rollout level, $K{=}20$ per cell). All four WM checkpoints are statistically flat regardless of the friction distribution seen at training time; both physics slopes are strictly negative.}
\label{tab:dr-flatness}
\begin{tabular}{@{}llrl@{}}
\toprule
Side & Checkpoint / policy & $\beta$ & 95\% CI \\
\midrule
WM      & seed 0 (fixed $\mu{=}1.0$)                  & $-0.009$ & $[-0.096, +0.082]$ \\
WM      & seed 1 (fixed $\mu{=}1.0$)                  & $+0.031$ & $[-0.072, +0.129]$ \\
WM      & seed 2 (fixed $\mu{=}1.0$)                  & $+0.038$ & $[-0.039, +0.123]$ \\
WM      & DR ($\mu \sim \mathcal{U}(0.1, 1.7)$)       & $-0.026$ & $[-0.123, +0.076]$ \\
\midrule
Physics & default policy (seed 0)                     & $-0.220$ & $[-0.301, -0.142]$ \\
Physics & DR policy                                   & $-0.114$ & $[-0.201, -0.024]$ \\
\bottomrule
\end{tabular}
\end{table}
 
\subsubsection{Per-step structure decomposition.}\label{app:perstep} 
\begin{figure*}[!ht]
  \centering
  \input{tex/perstep_compare_h64.tex}
  \caption{\textbf{Per-step iKCE structure.} Log-y. Physics (left) has discrete contact-event spikes whose positions shift with $\mu$. WM (right) shows a one- to two-step encoder-decoder transient followed by a smooth tail. The per-step structure is essentially constant across friction.}
  \label{fig:perstep}
\end{figure*}
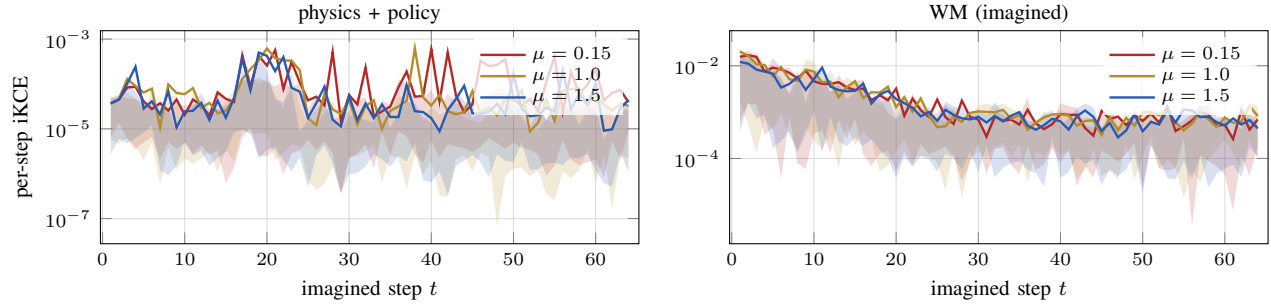

The integrated iKCE of Table~\ref{tab:h1} and Fig.~\ref{fig:headline} averages over the rollout horizon and so does not reveal whether the WM's imagined residual has the same temporal structure as physics. Fig.~\ref{fig:perstep} decomposes per-step iKCE at three friction values $\mu \in \{0.15, 1.0, 1.5\}$: physics exhibits sparse contact-event spikes whose positions shift with $\mu$ (consistent with footfall dynamics driving the kinematic-null residual), while the WM shows a one-to-two-step encoder-decoder transient followed by a smooth, friction-invariant tail. The two residuals are not merely different in magnitude (the H1 ratio) but qualitatively different in temporal structure, the WM's imagined rollouts do not reproduce the contact-event signature that defines the physics-side residual.
% Per-step decomposition --- shows physics has discrete contact spikes,
% WM has smooth tail (Control 2 in INITIAL_iKCE.md).

\subsubsection{Per-step structure under the actor-ablation checkpoint.}\label{app:perstep-actor}
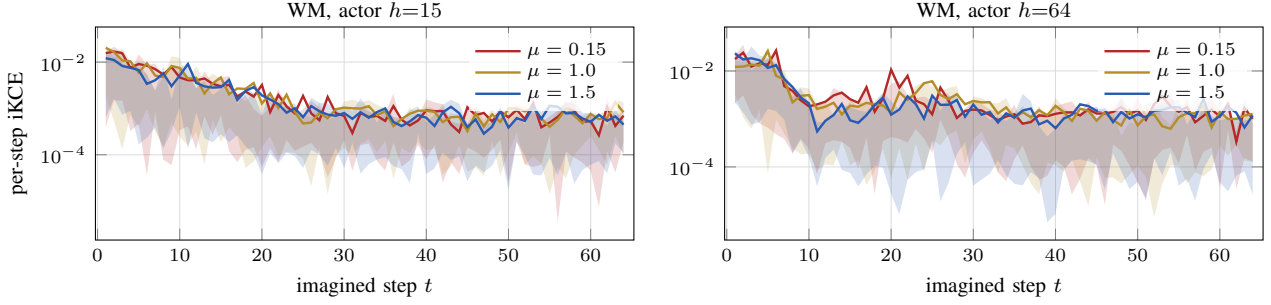
\begin{figure*}[!ht]
  \centering
  \input{tex/perstep_actor_ablation.tex}
  \caption{\textbf{Per-step actor ablation.} WM per-step iKCE at three friction values for both actor checkpoints. The
           transient-plus-smooth-tail structure is independent of the actor training horizon.}
  \label{fig:perstep_a64}
\end{figure*}

Figure~\ref{fig:perstep} reports the per-step decomposition under the default actor. Figure~\ref{fig:headline_a64} reports the integrated iKCE under the retrained $h{=}64$ actor. Figure~\ref{fig:perstep_a64} combines the two controls: per-step WM iKCE under both the default ($h{=}15$) and retrained ($h{=}64$) actor, at the same three friction values $\mu \in \{0.15, 1.0, 1.5\}$. The transient-plus-smooth-tail structure is unchanged across actor training horizons, ruling out the joint concern that the per-step signature reflects an actor-out-of-distribution artifact rather than a property of the WM's imagination.
% Per-step actor ablation --- direct h=15 vs h=64 actor comparison.

\subsubsection{Robustness to the kinematic-state choice.}\label{app:gait} 
\begin{figure*}[!ht]
  \centering
  \input{tex/headline_gait_h64.tex}
  \caption{\textbf{Gait DOFs view.} Same protocol as Fig.~\ref{fig:headline} but using the walker\_gait kinematic slice. Physics still shows low-$\mu$ elevation, WM is still flat, the finding is not an artifact of the $(z, v_z)$ identity view.}
  \label{fig:gait}
\end{figure*}
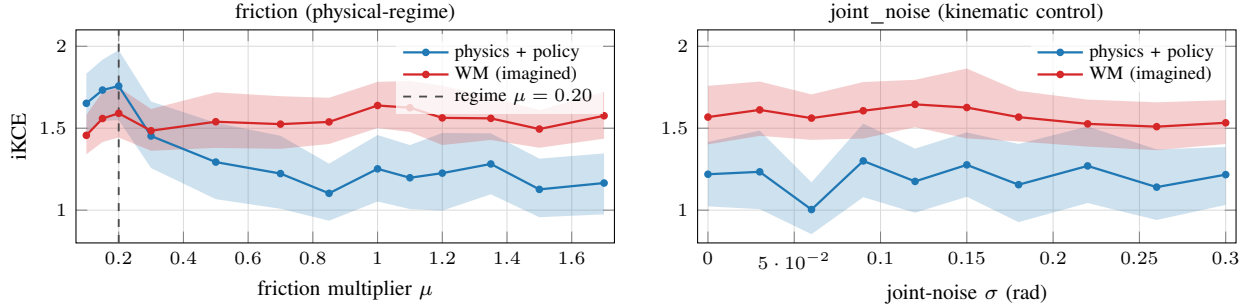

iKCE depends on a chosen kinematic state vector $\hat{x}_t$ (Definition~1). The main result uses the root-vertical-motion slice $(z, v_z)$. Figure~\ref{fig:gait} repeats the protocol with a richer representation, the walker's gait degrees of freedom. The qualitative pattern of Fig.~\ref{fig:headline} reappears: WM iKCE is flat across friction, while physics shows low-$\mu$ elevation, confirming that H2 reflects a property of the WM's imagination rather than the particular state slice used to evaluate it. This generalizes the diagnostic claim: any kinematic state extraction whose dynamics are sensitive to the perturbation axis can serve as a probe, with no special status accorded to the identity slice.
% Gait DOFs view --- alternative kinematic slice as sanity check.

\subsubsection{Joint-noise as a kinematic positive control.}\label{app:joint-noise}
The friction sweep is a dynamic perturbation (physically-grounded, regime-crossing). The joint-noise sweep is its kinematic counterpart (zero-mean Gaussian noise added to the joint-position channel of every observation before the WM encoder, with the physics state left unperturbed). A kinematic imaginer should respond to the joint-noise sweep. The perturbation directly modifies the kinematic state $\hat x_t$ that drives the extrapolation, while a dynamic imaginer should respond to the friction sweep. The right panel of Fig.~\ref{fig:headline} confirms that both channels respond to joint noise, ruling out the alternative explanation that the WM's iKCE is simply insensitive to all perturbations. The contrast (response to joint noise, non-response to friction) is the diagnostic signature named in the paper's title.

\subsection{Reproducibility}
\subsubsection{Code and data availability.}\label{app:code}
The diagnostic pipeline, trained checkpoints, perturbation-sweep CSVs, and PGFPlots figure sources for this paper are released at \url{https://github.com/TUM-AVS/iKCE}. The DreamerV3 implementation is the NM512 PyTorch port at commit \texttt{6ef8646}. The upstream algorithm is~\cite{hafner2023dreamerv3}. Checkpoints are released at \url{https://huggingface.co/fnc1901/ikce-walker-walk-artifacts}. All experiments ran on a single RTX~5090 (training: $\sim$24~h per checkpoint. The full perturbation sweep, including the actor-horizon ablation: $\sim$2~h).

\subsubsection{Implementation specifics.}\label{app:impl} 
For the identity view, the kinematic predictor is constant-velocity on the root vertical state: $\mathrm{kin}([z_t, \dot z_t]) = [z_t + \Delta t\,\dot z_t,\,\dot z_t]$ with $\Delta t = 25$\,ms (the DMC physics timestep). For the gait view, $\hat x_t$ stacks the unit-circle embedding $(\cos\theta_j, \sin\theta_j)$ of each walker joint angle, and $\mathrm{kin}(\cdot)$ applies one-step extrapolation per joint. Friction perturbations are applied by scaling the MuJoCo friction tuple of every geom by $\mu$ at episode reset (\texttt{model.geom\_friction[:, 0] *= mu}). Joint-noise perturbations add zero-mean Gaussian noise with standard deviation $\sigma$ (rad) to the joint-position channel of every observation before the WM encoder, with the physics state itself left unperturbed. For WM-imagined rollouts, we condition on the first 5 perturbed observations (encoder unroll), then roll out free imagination for the remaining $T-5$ steps. This matches the standard Dreamer evaluation protocol. All 95\% confidence intervals are computed using a percentile bootstrap with 1000 resamples across the K=20 rollouts per cell.

\end{document}

%% file: tex/styles.tex
\providecommand{\datadir}{.}

\definecolor{PhysicsColor}{RGB}{31,119,180}
\definecolor{WMColor}{RGB}{214,39,40}
\definecolor{RegimeColor}{RGB}{77,77,77}
\definecolor{LowMu}{RGB}{180,40,40}     % warm: low friction
\definecolor{MidMu}{RGB}{180,140,40}    % yellow-ish: baseline
\definecolor{HighMu}{RGB}{40,90,180}    % cool: high friction

\pgfplotsset{
  ikce/axis/.style={
    width=0.48\linewidth,
    height=4.4cm,
    grid=both,
    grid style={gray!18, line width=0.3pt},
    major grid style={gray!28},
    xlabel style={font=\footnotesize, yshift=1pt},
    ylabel style={font=\footnotesize, yshift=-2pt},
    tick label style={font=\scriptsize},
    title style={font=\footnotesize, yshift=-1.5ex},
    label style={font=\footnotesize},
    legend style={
      font=\scriptsize,
      draw=none,
      fill=white,
      fill opacity=0.78,
      text opacity=1,
      cells={anchor=west},
      inner sep=2pt,
      inner ysep=1.5pt,
      row sep=-2pt,
    },
    legend cell align=left,
    every axis plot/.append style={line width=0.85pt},
    enlarge x limits=0.02,
    scaled y ticks={base 10:0},  % use auto scaled-tick exponent; ylabel is short enough
    every y tick scale label/.style={
      at={(yticklabel cs:1.07)},   % push exponent further above plot, away from ylabel
      anchor=south west,
      font=\tiny,
      inner sep=0pt,
    },
  },
}

\pgfplotsset{
  ikce/ci/.style={fill opacity=0.22, draw=none},
  ikce/mean/.style={mark=*, mark size=1.0pt, line width=0.85pt},
  ikce/regime/.style={dashed, line width=0.7pt, color=RegimeColor},
}
\tikzset{
  ikce/ci/.style={fill opacity=0.22, draw=none},
  ikce/mean/.style={mark=*, mark size=1.0pt, line width=0.85pt},
  ikce/regime/.style={dashed, line width=0.7pt, color=RegimeColor},
}

%% file: tex/headline_template.tex
\newcommand{\drawiKCEheadline}[5]{%
  \begin{tikzpicture}
    \begin{groupplot}[
      group style={
        group size=2 by 1,
        horizontal sep=1.1cm,
      },
      ikce/axis,
    ]
      \nextgroupplot[
        xlabel={friction multiplier $\mu$},
        ylabel={iKCE},
        title={friction (physical-regime)},
        legend style={
          at={(0.98, 0.5)},
          anchor=east,
        },
        #5,
      ]
        \addplot [draw=none, forget plot, name path=phys_lo_F] table
          [col sep=comma, x=magnitude, y=ci_low]
          {\datadir/tex/data/#1/friction/ikce_summary.csv};
        \addplot [draw=none, forget plot, name path=phys_hi_F] table
          [col sep=comma, x=magnitude, y=ci_high]
          {\datadir/tex/data/#1/friction/ikce_summary.csv};
        \addplot [PhysicsColor, ikce/ci, forget plot]
          fill between [of=phys_lo_F and phys_hi_F];

        \addplot [draw=none, forget plot, name path=wm_lo_F] table
          [col sep=comma, x=magnitude, y=ci_low]
          {\datadir/tex/data/#2/friction/ikce_summary.csv};
        \addplot [draw=none, forget plot, name path=wm_hi_F] table
          [col sep=comma, x=magnitude, y=ci_high]
          {\datadir/tex/data/#2/friction/ikce_summary.csv};
        \addplot [WMColor, ikce/ci, forget plot]
          fill between [of=wm_lo_F and wm_hi_F];

        \addplot [PhysicsColor, ikce/mean] table
          [col sep=comma, x=magnitude, y=mean]
          {\datadir/tex/data/#1/friction/ikce_summary.csv};
        \addlegendentry{physics + policy}

        \addplot [WMColor, ikce/mean] table
          [col sep=comma, x=magnitude, y=mean]
          {\datadir/tex/data/#2/friction/ikce_summary.csv};
        \addlegendentry{WM (imagined)}

        \ifx\relax#4\relax\else
          \draw [ikce/regime]
            ({rel axis cs:0,0}-|{axis cs:#4,1}) --
            ({rel axis cs:0,1}-|{axis cs:#4,1});
          \addlegendimage{ikce/regime}
          \addlegendentry{regime $\mu = #4$}
        \fi

      \nextgroupplot[
        xlabel={joint-noise $\sigma$ (rad)},
        title={joint\_noise (kinematic control)},
        legend style={
          at={(0.98, 0.5)},
          anchor=east,
        },
        #5,
      ]
        \addplot [draw=none, forget plot, name path=phys_lo_J] table
          [col sep=comma, x=magnitude, y=ci_low]
          {\datadir/tex/data/#1/joint_noise/ikce_summary.csv};
        \addplot [draw=none, forget plot, name path=phys_hi_J] table
          [col sep=comma, x=magnitude, y=ci_high]
          {\datadir/tex/data/#1/joint_noise/ikce_summary.csv};
        \addplot [PhysicsColor, ikce/ci, forget plot]
          fill between [of=phys_lo_J and phys_hi_J];

        \addplot [draw=none, forget plot, name path=wm_lo_J] table
          [col sep=comma, x=magnitude, y=ci_low]
          {\datadir/tex/data/#2/joint_noise/ikce_summary.csv};
        \addplot [draw=none, forget plot, name path=wm_hi_J] table
          [col sep=comma, x=magnitude, y=ci_high]
          {\datadir/tex/data/#2/joint_noise/ikce_summary.csv};
        \addplot [WMColor, ikce/ci, forget plot]
          fill between [of=wm_lo_J and wm_hi_J];

        \addplot [PhysicsColor, ikce/mean] table
          [col sep=comma, x=magnitude, y=mean]
          {\datadir/tex/data/#1/joint_noise/ikce_summary.csv};
        \addlegendentry{physics + policy}

        \addplot [WMColor, ikce/mean] table
          [col sep=comma, x=magnitude, y=mean]
          {\datadir/tex/data/#2/joint_noise/ikce_summary.csv};
        \addlegendentry{WM (imagined)}

    \end{groupplot}
  \end{tikzpicture}%
}

%% file: tex/regime_curve.tex
% Regime collapse curve — episodic reward of the trained DreamerV3 walker
% policy across friction values. The empirical regime boundary (μ ≈ 0.20)
% is the friction at which the mean reward first drops below 50% of the
% μ=1.0 baseline. Used to set the dashed boundary line in the headline
% figure.

\begin{tikzpicture}
  \begin{axis}[
    ikce/axis,
    width=0.98\linewidth,
    height=5.2cm,
    xlabel={friction multiplier $\mu$},
    ylabel={episodic reward (0--1000)},
    xmin=0.05, xmax=1.55,
    ymin=0, ymax=900,
    legend pos=south east,
    legend style={
      font=\scriptsize,
      draw=none,
      fill=white,
      fill opacity=0.85,
      text opacity=1,
      inner sep=2pt,
      row sep=-2pt,
    },
    legend cell align=left,
  ]
    % 95% bootstrap CI band
    \addplot [draw=none, forget plot, name path=reglo] table
      [col sep=comma, x=mu, y=ci_low]
      {\datadir/tex/data/regime_boundary_friction.csv};
    \addplot [draw=none, forget plot, name path=reghi] table
      [col sep=comma, x=mu, y=ci_high]
      {\datadir/tex/data/regime_boundary_friction.csv};
    \addplot [PhysicsColor, ikce/ci, forget plot]
      fill between [of=reglo and reghi];

    % Mean reward curve
    \addplot [PhysicsColor, ikce/mean] table
      [col sep=comma, x=mu, y=mean_reward]
      {\datadir/tex/data/regime_boundary_friction.csv};
    \addlegendentry{trained policy reward}

    % Horizontal threshold line at 50% of μ=1.0 baseline (326).
    \addplot [dotted, color=RegimeColor, line width=0.7pt]
      coordinates {(0.05,326) (1.55,326)};
    \addlegendentry{50\% baseline}

    % Vertical regime boundary at μ = 0.20.
    \addplot [ikce/regime, color=WMColor]
      coordinates {(0.20,0) (0.20,900)};
    \addlegendentry{regime $\mu = 0.20$}
  \end{axis}
\end{tikzpicture}

%% file: tex/headline_identity_h64.tex
% Headline figure (paper's main result), identity view (z, vz), T=64.
% At this horizon the physics+policy curve develops a peak at the regime
% boundary μ=0.20 while the WM curve stays flat — the H2 signature.
\drawiKCEheadline
  {walker_walk_physics_policy_h64}
  {walker_walk_wm_h64}
  {{$(z, v_z)$, $T{=}64$}}
  {0.20}
  {ymode=log, ymin=2e-5, ymax=1e-2}

%% file: tex/horizon_sweep.tex
% Horizon-emergence sweep — slope of log(iKCE) on log(mu) at four
% measurement horizons T in {8, 16, 32, 64}, for the physics-side and
% WM-side friction sweeps. Shows quantitatively that the dynamic
% signature in physics emerges with horizon (slope magnitude grows,
% CI shifts off zero), while the WM remains statistically flat at
% every horizon.
%
% Data: tex/data/horizon_sweep_{physics,wm}.csv (each pre-filtered to
% one source for direct pgfplots consumption; the canonical
% long-format file horizon_sweep.csv lives alongside but is not used
% by this figure because pgfplots' `\addplot table[...]` cannot
% string-filter rows).
%
% Style: matches the regime_curve / headline figures —
% PhysicsColor / WMColor from styles.tex, ikce/axis for axis styling,
% ikce/ci for CI shading, ikce/mean for line+markers, fillbetween for
% the bands. A dashed zero-reference line uses the ikce/regime style.

\begin{tikzpicture}
  \begin{axis}[
    ikce/axis,
    width=0.98\linewidth,
    height=5.2cm,
    xmode=log,
    log basis x=2,
    xlabel={measurement horizon $T$ (steps)},
    ylabel={$\beta = \partial \log(\text{iKCE}) / \partial \log(\mu)$},
    xtick={8, 16, 32, 64},
    xticklabels={8, 16, 32, 64},
    xmin=6, xmax=80,
    ymin=-0.35, ymax=0.20,
    legend pos=south west,
    legend style={
      font=\scriptsize,
      draw=none,
      fill=white,
      fill opacity=0.85,
      text opacity=1,
      inner sep=2pt,
      row sep=-2pt,
    },
    legend cell align=left,
  ]
    % Zero reference — the H2 flatness target.
    \addplot [ikce/regime, forget plot]
      coordinates {(6,0) (80,0)};

    % ---------- Physics-side series ----------
    \addplot [draw=none, forget plot, name path=phys_lo] table
      [col sep=comma, x=horizon, y=ci_low]
      {\datadir/tex/data/horizon_sweep_physics.csv};
    \addplot [draw=none, forget plot, name path=phys_hi] table
      [col sep=comma, x=horizon, y=ci_high]
      {\datadir/tex/data/horizon_sweep_physics.csv};
    \addplot [PhysicsColor, ikce/ci, forget plot]
      fill between [of=phys_lo and phys_hi];

    \addplot [PhysicsColor, ikce/mean] table
      [col sep=comma, x=horizon, y=slope]
      {\datadir/tex/data/horizon_sweep_physics.csv};
    \addlegendentry{physics $+$ policy}

    % ---------- WM-side series ----------
    \addplot [draw=none, forget plot, name path=wm_lo] table
      [col sep=comma, x=horizon, y=ci_low]
      {\datadir/tex/data/horizon_sweep_wm.csv};
    \addplot [draw=none, forget plot, name path=wm_hi] table
      [col sep=comma, x=horizon, y=ci_high]
      {\datadir/tex/data/horizon_sweep_wm.csv};
    \addplot [WMColor, ikce/ci, forget plot]
      fill between [of=wm_lo and wm_hi];

    \addplot [WMColor, ikce/mean] table
      [col sep=comma, x=horizon, y=slope]
      {\datadir/tex/data/horizon_sweep_wm.csv};
    \addlegendentry{WM (imagined)}
  \end{axis}
\end{tikzpicture}

%% file: tex/headline_identity_a64.tex
% Ablation figure — h=64-trained actor checkpoint.
% Same protocol as headline_identity_h64.tex (T=64 measurement, identity
% view (z, vz)), but the DreamerV3 actor was trained with imag_horizon=64
% instead of the default 15. The retrained checkpoint reached ~520 mean
% reward (vs 955 for the h=15-actor default) — see CHECKPOINTS.md.
%
% Used in the supplementary / appendix as the actor-training-horizon
% ablation. The headline reading is that the WM curve is just as flat
% across friction here as in the main figure: friction-insensitivity
% does not depend on the actor's training horizon.
\drawiKCEheadline
  {walker_walk_physics_policy_a64}
  {walker_walk_wm_a64}
  {{$(z, v_z)$, $T{=}64$, actor $h{=}64$}}
  {0.20}
  {ymode=log, ymin=2e-5, ymax=1e-2}

%% file: tex/headline_identity_dr.tex
% DR-control headline figure — identity view (z, vz), T=64, under the
% domain-randomized DreamerV3 checkpoint (friction sampled per-episode
% from U(0.1, 1.7) at training time). The 4th macro argument is
% deliberately empty: under the DR policy there is no reward collapse
% at low friction (see docs/DR_RESULTS.md), so the empirical regime
% boundary at mu=0.20 used by the seed-0 headline figure does not
% apply here and the dashed line is omitted.
\drawiKCEheadline
  {walker_walk_physics_policy_h64_dr}
  {walker_walk_wm_h64_dr}
  {{$(z, v_z)$, $T{=}64$, DR policy}}
  {}
  {ymode=log, ymin=2e-5, ymax=1e-2}

%% file: tex/perstep_compare_h64.tex
% Per-step iKCE — physics + policy vs WM, T=64.
% Two-panel log-y comparison of per-step iKCE at three friction values:
% μ=0.15 (peak), μ=1.0 (baseline), μ=1.5 (over-friction).
%
% Physics shows discrete per-step spikes at contact events (with positions
% varying by μ). WM shows a step-1 encoder-decoder transient then a smooth
% decay tail — structurally constant across μ, consistent with kinematic
% imagination.

% Local helper: emits CI band + mean line for one (csv, color, label) triple.
% \def used so re-inputting this file doesn't error; the namespaced name
% avoids clashes with the document. If you input the file twice on
% purpose, the second \def silently wins (same body).
\def\ikcePerStepLine#1#2#3#4{%
  % #1 = unique name path tag, #2 = csv path, #3 = color, #4 = legend label
  \addplot [draw=none, forget plot, name path=#1lo] table
    [col sep=comma, x=step, y=q25] {#2};
  \addplot [draw=none, forget plot, name path=#1hi] table
    [col sep=comma, x=step, y=q75] {#2};
  \addplot [#3, fill opacity=0.18, draw=none, forget plot]
    fill between [of=#1lo and #1hi];
  \addplot [#3, line width=0.9pt] table
    [col sep=comma, x=step, y=mean] {#2};
  \addlegendentry{#4}%
}

\begin{tikzpicture}
  \begin{groupplot}[
    group style={
      group size=2 by 1,
      horizontal sep=1.2cm,
    },
    ikce/axis,
    width=0.48\linewidth,
    height=4.4cm,
    ymode=log,
    log basis y=10,
    xlabel={imagined step $t$},
    xmin=1, xmax=64,
  ]

    % ============== LEFT PANEL: physics + policy ==============
    \nextgroupplot[
      ylabel={per-step iKCE},
      title={physics + policy},
      legend pos=north east,
    ]
      \ikcePerStepLine{phys_low}{\datadir/tex/data/perstep_walker_walk_physics_policy_h64_friction_mu_0.15.csv}{LowMu}{$\mu = 0.15$}
      \ikcePerStepLine{phys_mid}{\datadir/tex/data/perstep_walker_walk_physics_policy_h64_friction_mu_1.csv}{MidMu}{$\mu = 1.0$}
      \ikcePerStepLine{phys_hi}{\datadir/tex/data/perstep_walker_walk_physics_policy_h64_friction_mu_1.5.csv}{HighMu}{$\mu = 1.5$}

    % ============== RIGHT PANEL: WM imagined ==============
    \nextgroupplot[
      title={WM (imagined)},
      legend pos=north east,
    ]
      \ikcePerStepLine{wm_low}{\datadir/tex/data/perstep_walker_walk_wm_h64_friction_mu_0.15.csv}{LowMu}{$\mu = 0.15$}
      \ikcePerStepLine{wm_mid}{\datadir/tex/data/perstep_walker_walk_wm_h64_friction_mu_1.csv}{MidMu}{$\mu = 1.0$}
      \ikcePerStepLine{wm_hi}{\datadir/tex/data/perstep_walker_walk_wm_h64_friction_mu_1.5.csv}{HighMu}{$\mu = 1.5$}

  \end{groupplot}
\end{tikzpicture}

%% file: tex/perstep_actor_ablation.tex
% Actor-training-horizon ablation, per-step iKCE.
%
% Compares the WM per-step iKCE structure on the h=15-actor (default)
% and h=64-actor (retrained) checkpoints, at three friction values.
% The strong claim from the main paper is that the WM produces a
% structurally-constant per-step pattern (transient + smooth tail)
% regardless of conditioning. The ablation shows this same pattern
% holds across actor training horizons: it's a property of WM
% imagination, not an artefact of one actor's training regime.
%
% Both panels show WM imagined rollouts at T=64; only the actor's
% imag_horizon training value differs.

\def\ikcePerStepLine#1#2#3#4{%
  \addplot [draw=none, forget plot, name path=#1lo] table
    [col sep=comma, x=step, y=q25] {#2};
  \addplot [draw=none, forget plot, name path=#1hi] table
    [col sep=comma, x=step, y=q75] {#2};
  \addplot [#3, fill opacity=0.18, draw=none, forget plot]
    fill between [of=#1lo and #1hi];
  \addplot [#3, line width=0.9pt] table
    [col sep=comma, x=step, y=mean] {#2};
  \addlegendentry{#4}%
}

\begin{tikzpicture}
  \begin{groupplot}[
    group style={
      group size=2 by 1,
      horizontal sep=1.2cm,
    },
    ikce/axis,
    width=0.48\linewidth,
    height=4.4cm,
    ymode=log,
    log basis y=10,
    xlabel={imagined step $t$},
    xmin=1, xmax=64,
    legend pos=north east,
  ]

    % ============== LEFT: WM, h=15 actor (paper's default) ==============
    \nextgroupplot[
      ylabel={per-step iKCE},
      title={WM, actor $h{=}15$},
    ]
      \ikcePerStepLine{wmh15_low}{\datadir/tex/data/perstep_walker_walk_wm_h64_friction_mu_0.15.csv}{LowMu}{$\mu = 0.15$}
      \ikcePerStepLine{wmh15_mid}{\datadir/tex/data/perstep_walker_walk_wm_h64_friction_mu_1.csv}{MidMu}{$\mu = 1.0$}
      \ikcePerStepLine{wmh15_hi}{\datadir/tex/data/perstep_walker_walk_wm_h64_friction_mu_1.5.csv}{HighMu}{$\mu = 1.5$}

    % ============== RIGHT: WM, h=64 actor (ablation) ==============
    \nextgroupplot[
      title={WM, actor $h{=}64$},
    ]
      \ikcePerStepLine{wmh64_low}{\datadir/tex/data/perstep_walker_walk_wm_a64_friction_mu_0.15.csv}{LowMu}{$\mu = 0.15$}
      \ikcePerStepLine{wmh64_mid}{\datadir/tex/data/perstep_walker_walk_wm_a64_friction_mu_1.csv}{MidMu}{$\mu = 1.0$}
      \ikcePerStepLine{wmh64_hi}{\datadir/tex/data/perstep_walker_walk_wm_a64_friction_mu_1.5.csv}{HighMu}{$\mu = 1.5$}

  \end{groupplot}
\end{tikzpicture}

%% file: tex/headline_gait_h64.tex
% Headline figure, gait-DOF view, T=64.
% Gait iKCE is dimensionless (atan2-based) and lives in [1.0, 1.8],
% so linear-y reads better than the log-y used for the (z, v_z) view.
\drawiKCEheadline
  {walker_walk_physics_policy_gait_h64}
  {walker_walk_wm_gait_h64}
  {{walker gait DOFs, $T{=}64$}}
  {0.20}
  {ymode=linear, ymin=0.8, ymax=2.1,
   legend style={at={(0.98, 0.98)}, anchor=north east}}